%% file: ms.tex
\documentclass{article}

\usepackage[final]{nips_2017}
\usepackage[colorlinks=true]{hyperref}
\usepackage[pdftex]{graphicx}
\graphicspath{figures/}
\usepackage{url}
\usepackage{amsmath}
\usepackage{amssymb}
\usepackage{wrapfig}
\usepackage{float}
\usepackage{mathtools}
\usepackage{enumitem}
\usepackage{tabu}

\usepackage{color,xcolor}
\usepackage{xspace}
\usepackage{booktabs}
\usepackage{titlesec}
\input{macros}
\hypersetup{citecolor=gray}
\hypersetup{urlcolor=gray}
\hypersetup{linkcolor=red}

\begin{document}

\title{Self-Supervised Intrinsic Image Decomposition}

\author{Michael Janner\\
MIT \\
{\tt\small janner@mit.edu}
\and
\textbf{Jiajun Wu}\\
MIT \\
{\tt\small jiajunwu@mit.edu}
\and
\textbf{Tejas D. Kulkarni} \\
DeepMind \\
{\tt\small tejasdkulkarni@gmail.com}
\AND
\textbf{Ilker Yildirim} \\
MIT \\
{\tt\small ilkery@mit.edu}
\and
\textbf{Joshua B. Tenenbaum} \\
MIT \\
{\tt\small jbt@mit.edu}
}

\maketitle


\input{text/abstract}

\input{text/intro}
\input{text/related_work}
\input{text/model}

\input{text/experiments}
\input{text/conclusion}


\subsection*{Acknowledgements}

This work is supported by ONR MURI N00014-16-1-2007, the Center for Brain, Minds and Machines (NSF \#1231216), Toyota Research Institute, and Samsung.

{\small
\bibliographystyle{plainnat}
\bibliography{intrinsic}
}
\end{document}

%% file: macros.tex
\newcolumntype{L}[1]{>{\raggedright\let\newline\\\arraybackslash\hspace{0pt}}m{#1}}
\newcolumntype{C}[1]{>{\centering\let\newline\\\arraybackslash\hspace{0pt}}m{#1}}
\newcolumntype{R}[1]{>{\raggedleft\let\newline\\\arraybackslash\hspace{0pt}}m{#1}}

\newcommand{\eqn}[1]{Equation~\ref{#1}}

\newcommand{\ignore}[1]{}

\makeatletter
\DeclareRobustCommand\onedot{\futurelet\@let@token\@onedot}
\def\@onedot{\ifx\@let@token.\else.\null\fi\xspace}

\makeatother

\definecolor{MyDarkBlue}{rgb}{0,0.08,1}
\definecolor{MyDarkGreen}{rgb}{0.02,0.6,0.02}
\definecolor{MyDarkRed}{rgb}{0.8,0.02,0.02}
\definecolor{MyDarkOrange}{rgb}{0.40,0.2,0.02}
\definecolor{MyPurple}{RGB}{111,0,255}
\definecolor{MyRed}{rgb}{1.0,0.0,0.0}
\definecolor{MyGold}{rgb}{0.75,0.6,0.12}
\definecolor{MyDarkgray}{rgb}{0.66, 0.66, 0.66}


%% file: text/abstract.tex
\begin{abstract}
Intrinsic decomposition from a single image is a highly challenging task, due to its inherent ambiguity and the scarcity of training data. In contrast to traditional fully supervised learning approaches, in this paper we propose learning intrinsic image decomposition by explaining the input image. Our model, the Rendered Intrinsics Network (\emph{RIN}), joins together an image decomposition pipeline, which predicts reflectance, shape, and lighting conditions given a single image, with a recombination function, a learned shading model used to recompose the original input based off of intrinsic image predictions. Our network can then use unsupervised reconstruction error as an additional signal to improve its intermediate representations. This allows large-scale unlabeled data to be useful during training, and also enables transferring learned knowledge to images of unseen object categories, lighting conditions, and shapes. Extensive experiments demonstrate that our method performs well on both intrinsic image decomposition and knowledge transfer.


\end{abstract}

%% file: text/intro.tex
\section{Introduction}
\label{sec:intro}

There has been remarkable progress in computer vision, particularly for answering questions such as \textit{``what is where?"} given raw images. This progress has been possible due to large labeled training sets and representation learning techniques such as convolutional neural networks \citep{lecun2015deep}. However, the general problem of visual scene understanding will require algorithms that extract not only object identities and locations, but also their shape, reflectance, and interactions with light. Intuitively disentangling the contributions from these three components, or \emph{intrinsic images}, is a major triumph of human vision and perception. Conferring this type of intuition to an algorithm, though, has proven a difficult task, constituting a major open problem in computer vision.  

\begin{figure}[t]
    \centering
    \includegraphics[width=1.0\linewidth]{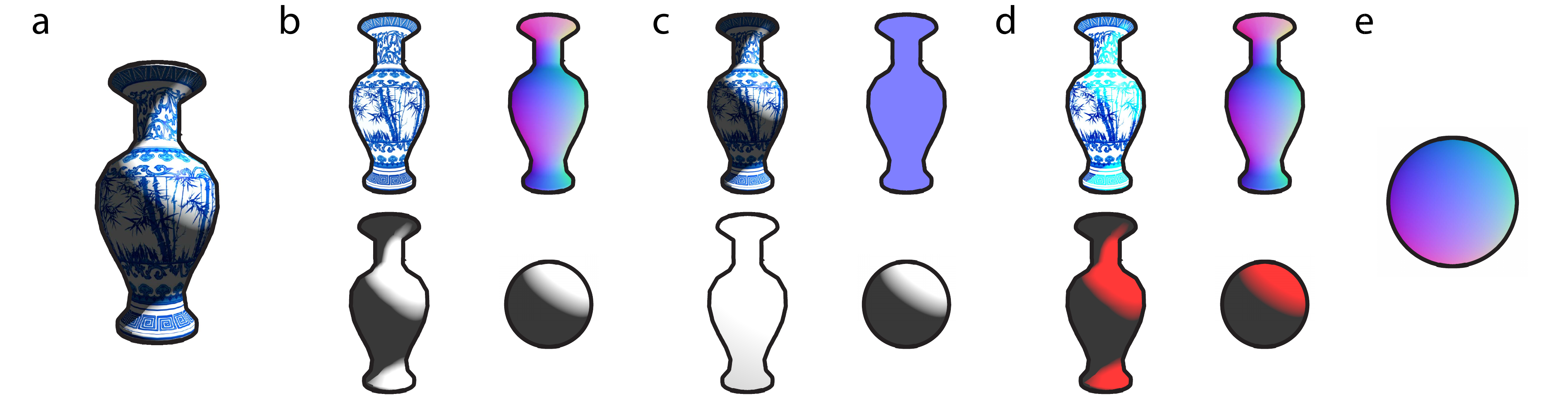}
    \vspace{10pt}
    \caption{A porcelain vase (a) along with three predictions (b-d) for its underlying intrinsic images. The set in (c) assumes the contribution from shading is negligible by predicting a completely flat rather than rounded shape. The reflectance is therefore indistinguishable from the observed image. The set in (d) includes the correct shape but assumes red lighting and a much brighter blue color in the regions affected by shading. While the decomposition in (b) is much more inuitively pleasing than either of these alternatives, all of these options are valid in that they combine to exactly form the observed vase. (e) shows a sphere with our visualized normals map as a shape reference.}
    \label{fig:underconstrained}
\end{figure}

This problem is challenging in particular because it is fundamentally underconstrained. Consider the porcelain vase in Figure~\ref{fig:underconstrained}a. Most individuals would have no difficulty identifying the true colors and shape of the vase, along with estimating the lighting conditions and the resultant shading on the object, as those shown in \ref{fig:underconstrained}b. However, the alternatives in \ref{fig:underconstrained}c, which posits a flat shape, and \ref{fig:underconstrained}d, with unnatural red lighting, are entirely consistent in that they compose to form the correct observed vase in \ref{fig:underconstrained}a. 


The task of finding appropriate intrinsic images for an object is then not a question of simply finding a valid answer, as there are countless factorizations that would be equivalent in terms of their rendered combination, but rather of finding the most probable answer. Roughly speaking, there are two methods of tackling such a problem: a model must either (1) employ handcrafted priors on the reflectance, shape, and lighting conditions found in the natural world in order to assign probabilities to intrinsic image proposals or (2) have access to a library of ground truth intrinsic images and their corresponding composite images. 

Unfortunately, there are limitations to both methods. Although there has been success with the first route in the past \citep{sirfs}, strong priors are often difficult to hand-tune in a generally useful fashion. On the other hand, requiring access to complete, high quality ground truth intrinsic images for real world scenes is also limiting, as creating such a training set requires an enormous amount of human effort and millions of crowd-sourced annotations \citep{crowdsourced}. 

In this paper, we propose a deep structured autoencoder, the Rendered Instrinsics Network (\emph{RIN}), that disentangles intrinsic image representations and uses them to reconstruct the input. The decomposition model consists of a shared convolutional encoder for the observation and three separate decoders for the reflectance, shape, and lighting. The shape and lighting predictions are used to train a differentiable shading function. The output of the shader is combined with the reflectance prediction to reproduce the observation. The minimal structure imposed in the model -- namely, that intrinsic images provide a natural way of disentangling real images and that they provide enough information to be used as input to a graphics engine -- makes RIN act as an autoencoder with useful intermediate representations.

The structure of RIN also exploits two natural sources of supervision: one applied to the intermediate representations themselves, and the other to the reconstructed image. This provides a way for RIN to improve its representations with unlabeled data. By avoiding the need for intrinsic image labels for all images in the dataset, RIN can adapt to new types of inputs even in the absence of ground truth data. We demonstrate the utility of this approach in three transfer experiments. RIN is first trained on a simple set of five geometric primitives in a supervised manner and then transferred to common computer vision test objects.  Next, RIN is trained on a dataset with a skewed underlying lighting distribution and fills in the missing lighting conditions on the basis of unlabeled observations. Finally, RIN is trained on a single Shapenet category and then transferred to a separate, highly dissimilar category. 

Our contributions are three-fold. First, we propose a novel formulation for intrinsic image decomposition, incorporating a differentiable, unsupervised reconstruction loss into the loop. Second, we instantiate this approach with the RIN, a new model that uses convolutional neural networks for both intrinsic image prediction and recombination via a learned shading function.
This is also the first work to apply deep learning to the full decomposition into reflectance, shape, lights, and shading, as prior work has focused on the reflectance-shading decomposition. Finally, we show that RIN can make use of unlabeled data to improve its intermediate intrinsic image representations and transfer knowledge to new objects unseen during training.

%% file: text/related_work.tex
\begin{figure*}[t]
    \centering
    \includegraphics[width=1.0\linewidth]{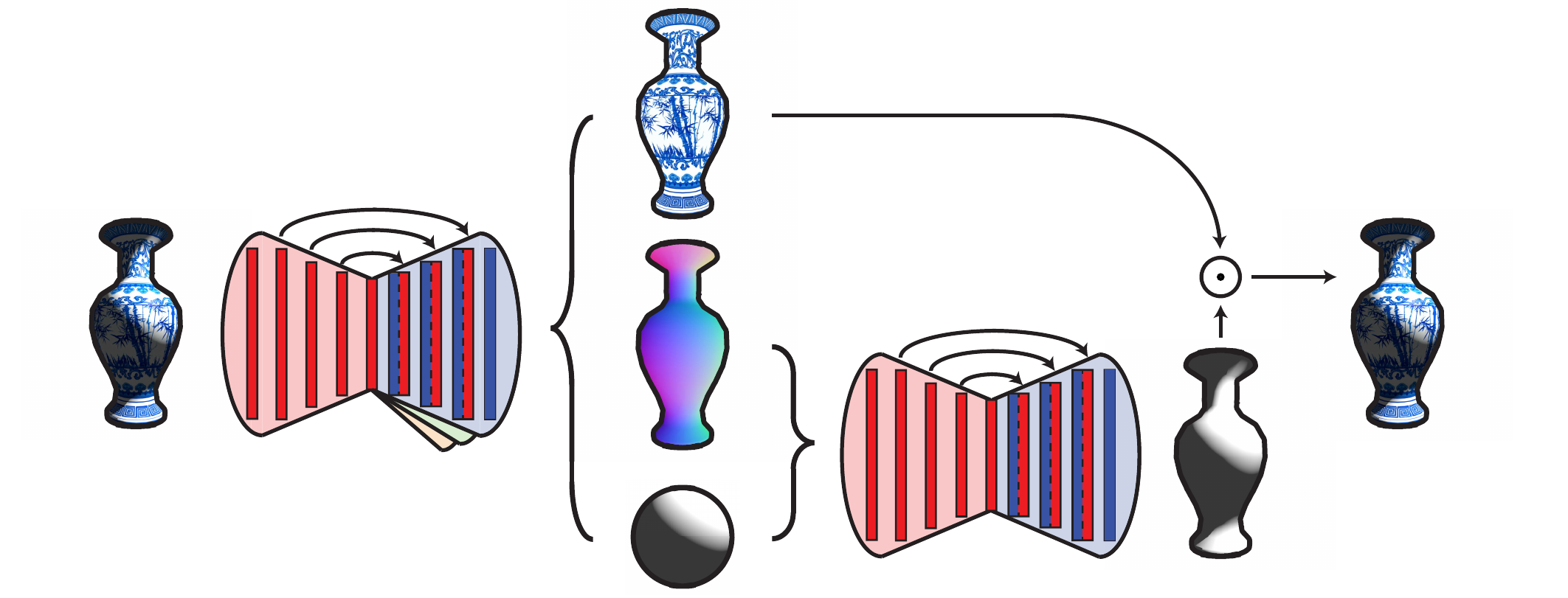}
    \caption{RIN contains two convolutional encoder-decoders, one used for predicting the intrinsic images from an input and another for predicting the shading stemming from a light source applied on a shape. The two networks together function as a larger structured autoencoder, forcing a specific type of intermediate representation in order to reconstruct the input image.}
    \label{fig:schematic}
\end{figure*}

\section{Related Work}
\label{sec:related}

Intrinsic images were introduced by Barrow and Tenenbaum as useful mid-level scene descriptors~\citep{barrow_tenenbaum}. The model posits that an image can be expressed as the pointwise product between contributions from the true colors of an object, or its reflectance, and contributions from the shading on that object:
\begin{equation} 
\label{eq:decomposition}
\text{image } I= \text{reflectance } R \cdot \text{shading } S
\end{equation}
Decomposing one step further, the shading is expressed as some function of an object's shape and the ambient lighting conditions. The exact nature of this shading function varies by implementation.

Early work on intrinsic image decomposition was based on insights from Land's Retinex Theory~\citep{retinex}. \cite{horn} separated images into true colors and shading using the assumption that large image gradients tend to correspond to reflectance changes and small gradients to lighting changes. While this assumption works well for a hypothetical \emph{Mondrian World} of flat colors, it does not always hold for natural images. In particular, \cite{sequences} found that this model of reflectance and lighting is rarely true for outdoor scenes. 

More recently, \cite{sirfs} developed an iterative algorithm called SIRFS that maximizes the likelihood of intrinsic image proposals under priors derived from regularities in natural images. SIRFS proposes shape and lighting estimates and combines them via a spherical harmonics renderer to produce a shading image. \cite{lombardi_cvpr_2012,lombardi_tpami_2016} and \cite{oxholm_tpami_2016} proposed a Bayesian formulation of such an optimization procedure, also formulating priors based on the distribution of material properties and the physics of lighting in the real world. Researchers have also explored reconstructing full 3D shapes through intrinsic images by making use of richer generative models~\citep{category_specific,marrnet}.

\cite{dln} combined Lambertian reflectance assumptions with Deep Belief Networks to learn a prior over the reflectance of greyscale images and applied their \emph{Deep Lambertian Network} to one-shot face recognition. \cite{narihari_lightness} applied deep learning to intrinsic images first using human judgments on real images and later in the context of animated movie frames~\citep{direct_intrinsics}. \cite{deep_reflectance} and \cite{deep_illumination} also used convolutional neural networks to estimate reflectance maps and illumination parameters, respectively, in unconstrained outdoor settings. 

\cite{innamorati} generalized the intrinsic image decomposition by considering the contributions of specularity and occlusion in a direction-dependent model. \cite{non_lambertian} found improved performance in the full decomposition by incorporating skip layer connections \citep{residual_learning} in the network architecture, which were used to generate much crisper images. Our work can be seen as a further extension of these models which aims to relax the need for a complete set of ground truth data by modeling the image combination process, as in \cite{deep_shading}.

Incorporating a domain-specific decoder to reconstruct input images has been explored by Hinton \emph{et al.} in their \emph{transforming autoencoders} \citep{hinton_autoencoder}, which also learned natural representations of images in use by the vision community. Our work differs in the type of representation in question, namely images rather than descriptors like affine transformations or positions. \cite{dcign} were also interested in learning disentangled representations in an autoencoder, which they achieved by selective gradient updates during training. Similarly, \cite{infogan} showed that a mutual information objective could drive disentanglement of a deep network's intermediate representation. 

%% file: text/model.tex
\section{Model}
\label{sec:problem}

\subsection{Use of Reconstruction}
RIN differs most strongly with past work in its use of the reconstructed input. Other approaches have fallen into roughly two groups in this regard:
\begin{enumerate}[leftmargin=0.5cm]
\item Those that solve for one of the intrinsic images to match the observed image. SIRFS, for example, predicts shading and then solves equation~\ref{eq:decomposition} for reflectance given its prediction and the input~\citep{sirfs}. This ensures that the intrinsic image estimations combine to form exactly the observed image, but also deprives the model of any reconstruction error. 
\item Data-driven techniques that rely solely on ground truth labelings~\citep{direct_intrinsics, non_lambertian}. These approaches assume access to ground truth labels for all inputs and do not explicitly model the reconstruction of the input image based on intrinsic image predictions. 
\end{enumerate}

Making use of the reconstruction for this task has been previously unexplored because such an error signal can be difficult to interpret. Just as the erroneous intrinsic images in Fig~\ref{fig:underconstrained}c-d combine to reconstruct the input exactly, one cannot assume that low reconstruction error implies accurate intrinsic images. An even simpler degenerate solution that yields zero reconstruction error is: 
\begin{equation} 
\label{eq:degenerate}
\hat{R}= I \quad\text{and}\quad \hat{S} = \mathbf{1}\mathbf{1}^T,
\end{equation}
where $\hat{S}$ is the all-ones matrix. It is necessary to further constrain the predictions such that the model does not converge to such explanations. 

\begin{figure}[t]
    \centering
    \includegraphics[width=1.0\linewidth]{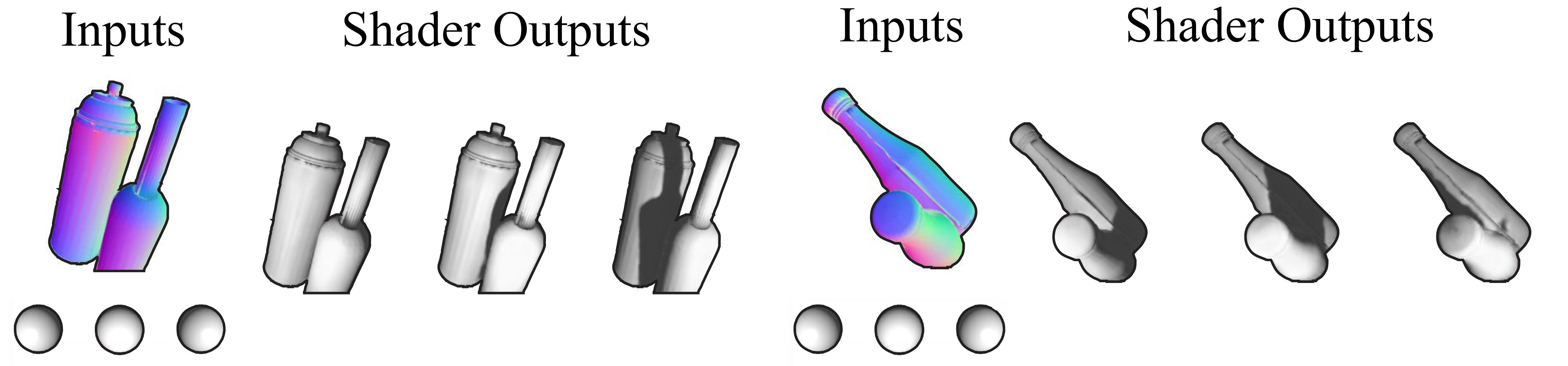}
    \caption{In contrast to simple Lambertian shading techniques, our learned shading model can handle shadows cast between objects. Inputs to the shader are shape and lighting parameter pairs.}
    \label{fig:shadows}
\end{figure}

\subsection{Shading Engine}
RIN decomposes an observation into reflectance, shape, and lighting conditions. As opposed to models which estimate only reflectance and shading, which may make direct use of \eqn{eq:decomposition} to generate a reconstruction, we must employ a function that transforms our shape and lighting predictions into a shading estimate. Linear Lambertian assumptions could reduce such a function to a straightforward dot product, but would produce a shading function incapable of modeling lighting conditions that drastically change across an image or ray-tracing for the purposes of casting shadows.

Instead, we opt to learn a shading model. Such a model is not limited in the way that a pre-defined shading function would be, as evidenced by shadows cast between objects in Fig~\ref{fig:shadows}. Learning a shader also has the benefit of allowing for different representations of lighting conditions. In our experiments, lights are defined by a position in three-dimensional space and a magnitude, but alternate representations such as the radius, orientation, and color of a spotlight could be just as easily adopted. For work that employs the shading engine from SIRFS \citep{sirfs} instead of learning a shader in a similar disentanglement context, see \cite{shu2017face}. The SIRFS engine represents lights as spherical harmonics coefficient vectors. 

\begin{figure}[t]
    \centering
    \includegraphics[width=1.0\linewidth]{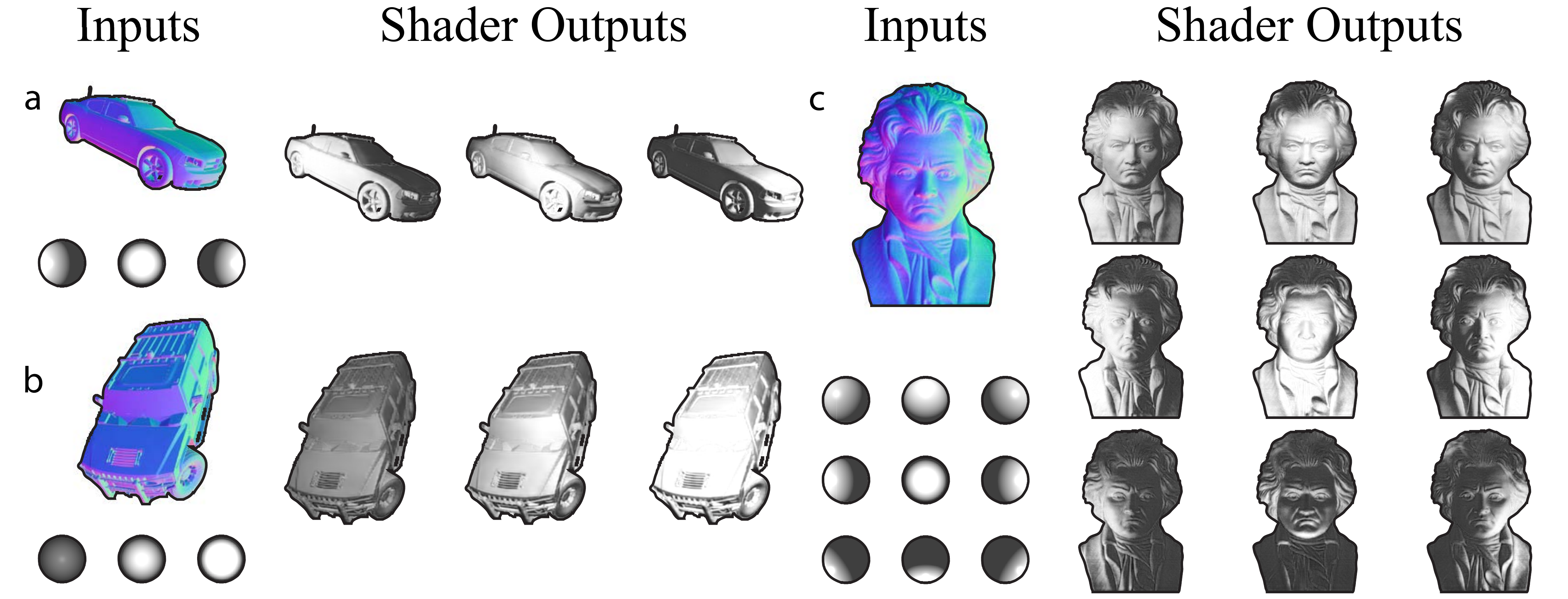}
    \caption{Our shading model's outputs after training only on synthetic car models from the ShapeNet dataset~\citep{shapenet}. (a) shows the effect of panning the light horizontally and (b) shows the effect of changing the intensity of the light. The input lights are visualized by rendering them onto a sphere. Even though the shader was trained only on synthetic data, it generalizes well to real shapes with no further training. The shape input to (c) is an estimated normals map of a Beethoven bust~\citep{beethoven}.}
    \label{fig:shader}
\end{figure}

\subsection{Architecture}
Our model consists of two convolutional encoder-decoder networks, the first of which predicts intrinsic images from an observed image, and the second of which approximates the shading process of a rendering engine. Both networks employ mirror-link connections introduced by \cite{non_lambertian}, which connect layers of the encoder and decoder of the same size. These connections yield sharper results than the blurred outputs characteristic of many deconvolutional models. 

The first network has a single encoder for the observation and three separate decoders for the reflectance, lighting, and shape. Unlike \cite{non_lambertian}, we do not link layers between the decoders so that it is possible to update the weights of one of the decoders without substantially affecting the others, as is useful in the transfer learning experiments. The encoder has 5 convolutional layers with \{16, 32, 64, 128, 256\} filters of size 3$\times$3 and stride of 2. Batch normalization~\citep{batchnorm} and ReLU activation are applied after every convolutional layer. The layers in the reflectance and shape decoders have the same number of features as the encoder but in reverse order plus a final layer with 3 output channels. Spatial upsampling is applied after the convolutional layers in the decoders. The lighting decoder is a simple linear layer with an output dimension of four (corresponding to a position in three-dimensional space and an intensity of the light).

The shape is passed as input to the shading encoder directly. The lighting estimate is passed to a fully-connected layer with output dimensionality matching that of the shading encoder's output, which is concatenated to the encoded shading representation. The shading decoder architecture is the same as that of the first network. The final component of RIN, with no learnable parameters, is a componentwise multiplication between the output of the shading network and the predicted reflectance.

%% file: text/experiments.tex
\begin{figure}[t]
    \centering
    \includegraphics[width=1.0\linewidth]{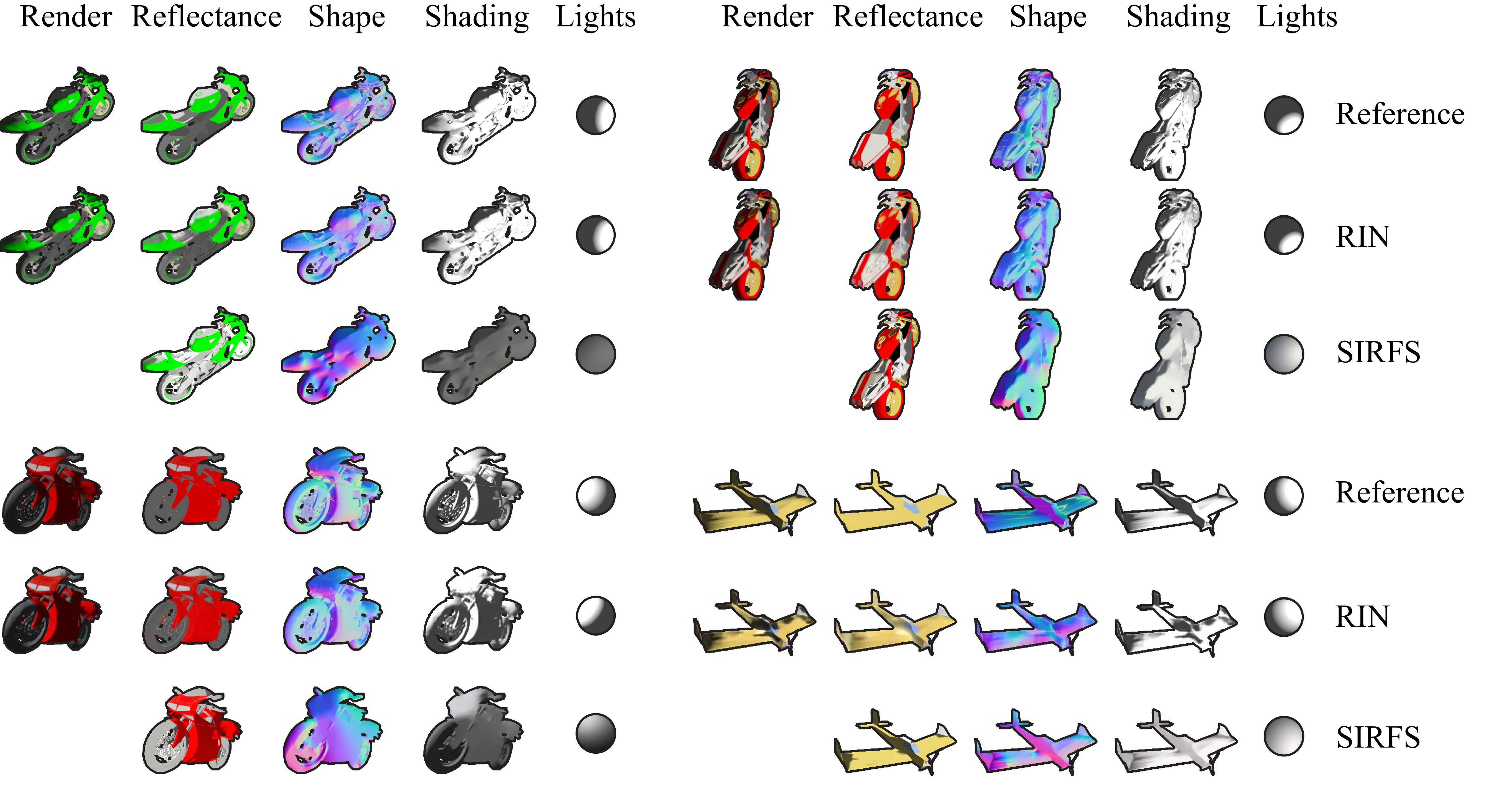}
    \caption{Intrinsic image prediction from our model on objects from the training category (motorbikes) as well as an example from outside this category (an airplane). The quality of the airplane intrinsic images is significantly lower, which is reflected in the reconstruction (labeled "Render" in the RIN rows). This allows reconstruction to drive the improvement of the intermediate intrinsic image representations. Predictions from SIRFS are shown for comparison. Note that the reflectance in SIRFS is defined based on the difference between the observation and shading prediction, so there is not an analogous reconstruction.}
    \label{fig:supervised}
\end{figure}

\begin{table}[t]
    \centering
    \begin{tabular}{lcccccc}
        \toprule
        & \multicolumn{3}{c}{\textbf{Motorbike (Train)}} & \multicolumn{3}{c}{\textbf{Airplane (Transfer)}} \\
        \cmidrule(lr){2-4}\cmidrule(lr){5-7}
        & Reflectance & Shape & Lights & Reflectance & Shape & Lights \\
        \midrule
        RIN &  0.0021 & 0.0044 & 0.1398 & 0.0042 & 0.0119 & 0.4873 \\
        SIRFS & 0.0059 & 0.0094 & -- & 0.0054 & 0.0080 & -- \\
        \bottomrule
    \end{tabular}
    \label{tbl:supervised}
    \vspace{0.5cm}
    \caption{MSE of our model and SIRFS on a test set of ShapeNet motorbikes, the category used to train RIN, and airplanes, a held-out class. The lighting representation of SIRFS (a vector with 27 components) is sufficiently different from that of our model that we do not attempt to compare performance here directly. Instead, see the visualization of lights in Fig~\ref{fig:supervised}.}
\end{table}

\section{Experiments}
\label{sec:exp}
RIN makes use of unlabeled data by comparing its reconstruction to the original input image. Because our shading model is fully differentiable, as opposed to most shaders that involve ray-tracing, the reconstruction error may be backpropagated to the intrinsic image predictions and optimized via a standard coordinate ascent algorithm. RIN has one shared encoder for the intrinsic images but three separate decoders, so the appropriate decoder can be updated while the others are held fixed. 

In the following experiments, we first train RIN (including the shading model) on a dataset with ground truth labels for intrinsic images. This is treated as a standard supervised learning problem using mean squared error on the intrinsic image predictions as a loss. The model is then trained further on an additional set of \emph{unlabeled} data using only reconstruction loss as an error signal. We refer to this as the self-supervised transfer. For both modes of learning, we optimize using Adam~\citep{adam}. 

During transfer, one half of a minibatch will consist of the unlabeled transfer data the other half will come from the labeled data. This ensures that the representations do not shift too far from those learned during the initial supervised phase, as the underconstrained nature of the problem can drive the model to degenerate solutions. When evaluating our model on test data, we use the outputs of the three decoders and the learned shader directly; we do not enforce that the predictions must explain the input exactly.

Below, we demonstrate that our model can effectively transfer to different shapes, lighting conditions, and object categories without ground truth intrinsic images. However, for this unsupervised transfer to yield benefits, there must be a sufficient number of examples of the new, unlabeled data. For example, the MIT Intrinsic Images dataset~\citep{mit_intrinsic}, containing twenty real-world images, is not large enough for the unsupervised learning to affect the representations of our model. In the absence of any unsupervised training, our model is similar to that of \cite{non_lambertian} adapted to predict the full set of intrinsic images.

\begin{figure}[t]
    \centering
    \includegraphics[width=1.0\linewidth]{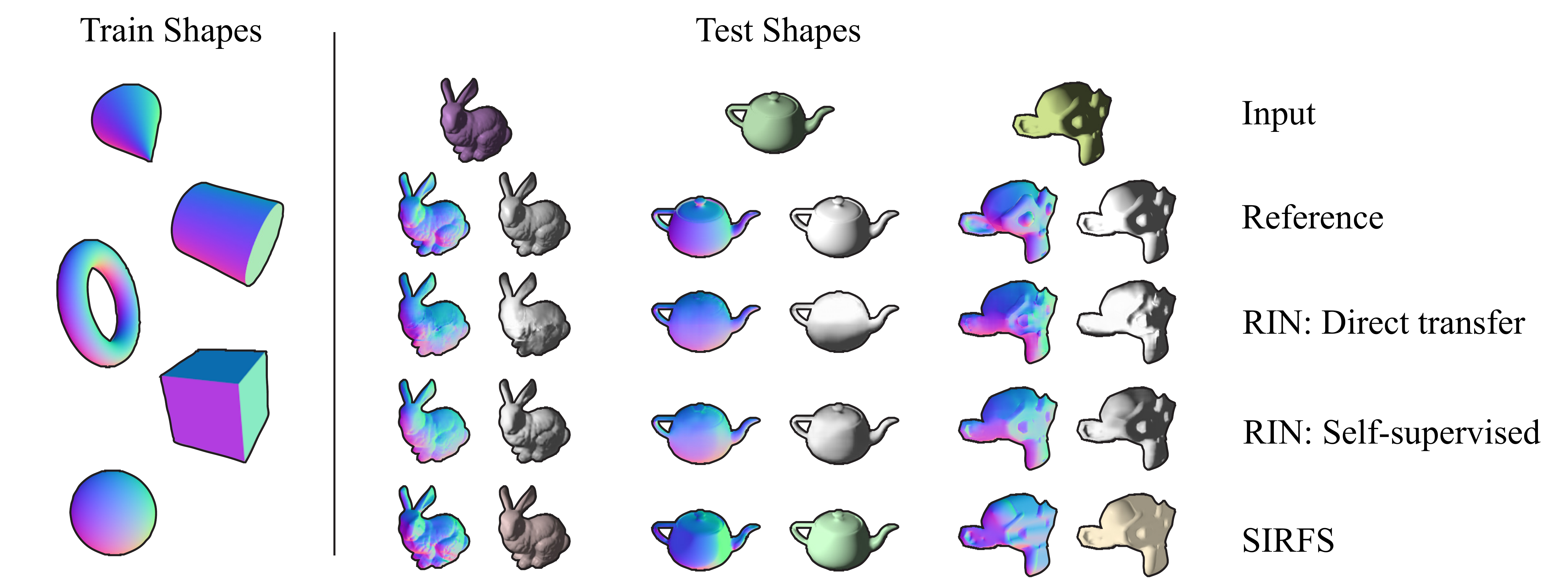}
    \caption{Predictions of RIN before ("Direct transfer") and after ("Self-supervised") it adapts to new shapes on the basis of unlabeled data.}
    \label{fig:gen_shapes}
\end{figure}

\begin{table}[t]
    \centering
    \begin{tabular}{lcccccc}
        \toprule
        & \multicolumn{2}{c}{\textbf{Stanford Bunny}} & \multicolumn{2}{c}{\textbf{Utah Teapot}} & \multicolumn{2}{c}{\textbf{Blender Suzanne}} \\
        \cmidrule(lr){2-3}\cmidrule(lr){4-5}\cmidrule(lr){6-7}
        & Shape & Shading & Shape & Shading & Shape & Shading \\
        \midrule
        Direct transfer &  0.074 & 0.071 & 0.036 & 0.043 & 0.086 & 0.104\\
        Self-supervised & 0.048 & 0.005 & 0.029 & 0.003 & 0.058 & 0.007 \\
        \bottomrule
    \end{tabular}
    \vspace{.5cm}
    \caption{MSE of RIN trained on five geometric primitives before and after self-supervised learning of more complicated shapes.}
    \label{tbl:gen_shapes_}
\end{table}

\subsection{Supervised training}
\label{sec:supervised}

\textbf{Data} \hspace{0.35cm} The majority of data was generated from ShapeNet~\citep{shapenet} objects rendered in Blender. For the labeled datasets, the rendered composite images were accompanied by the object's reflectance, a map of the surface normals at each point, and the parameters of the lamp used to light the scene. Surface normals are visualized by mapping the XYZ components of normals to appropriate RGB ranges. For the following supervised learning experiments, we used a dataset size of 40,000 images.

\textbf{Intrinsic image decomposition} \hspace{0.35cm} The model in Fig~\ref{fig:supervised} was trained on ShapeNet motorbikes. Although it accurately predicts the intrinsic images of the train class, its performance drops when tested on other classes. In particular, the shape predictions suffer the most, as they are the most dissimilar from anything seen in the training set. Crucially, the poor intrinsic image predictions are reflected in the reconstruction of the input image. This motivates the use of reconstruction error to drive improvement of intrinsic images when there is no ground truth data. 

\textbf{Shading model} \hspace{0.35cm} In contrast with the intrinsic image decomposition, shading prediction generalized well outside of the training set. The shader was trained on the shapes and lights from the same set of rendered synthetic cars as above. Even though this represents only a narrow distribution over shapes, we found that the shader produced plausible predictions for even real-world objects (Fig~\ref{fig:shader}). Because the shader generalized without any further effort, its parameters were never updated during self-supervised training. Freezing the parameters of the shader prevents our model from producing nonsensical shading images.  

\subsection{Shape transfer}

\textbf{Data} \hspace{0.35cm} 
We generated a dataset of five shape primitives (cubes, spheres, cones, cylinders, and toruses) viewed at random orientations using the Blender rendering engine. These images are used for supervised training. 
Three common reference shapes (Stanford bunny, Utah teapot, and Blender's Suzanne) are used as the unlabeled transfer class. To isolate the effects of shape mismatch in the labeled versus unlabeled data, all eight shapes were rendered with random monochromatic materials and a uniform distribution over lighting positions within a contained region of space in front of the object. The datasets consisted of each shape rendered with 500 different colors, with each colored shape being viewed at 10 orientations.   

\textbf{Results} \hspace{0.35cm} By only updating weights for the shape decoder during self-supervised transfer, the predictions for held-out shapes improves by 29\% (averaged across the three shapes). Because a shape only affects a rendered image via shading, the improvement in shapes comes alongside an improvement in shading predictions as well. Shape-specific results are given in Table~\ref{tbl:gen_shapes_} and visualized in Fig~\ref{fig:gen_shapes}.

\begin{figure}[t]
    \centering
    \includegraphics[width=\linewidth]{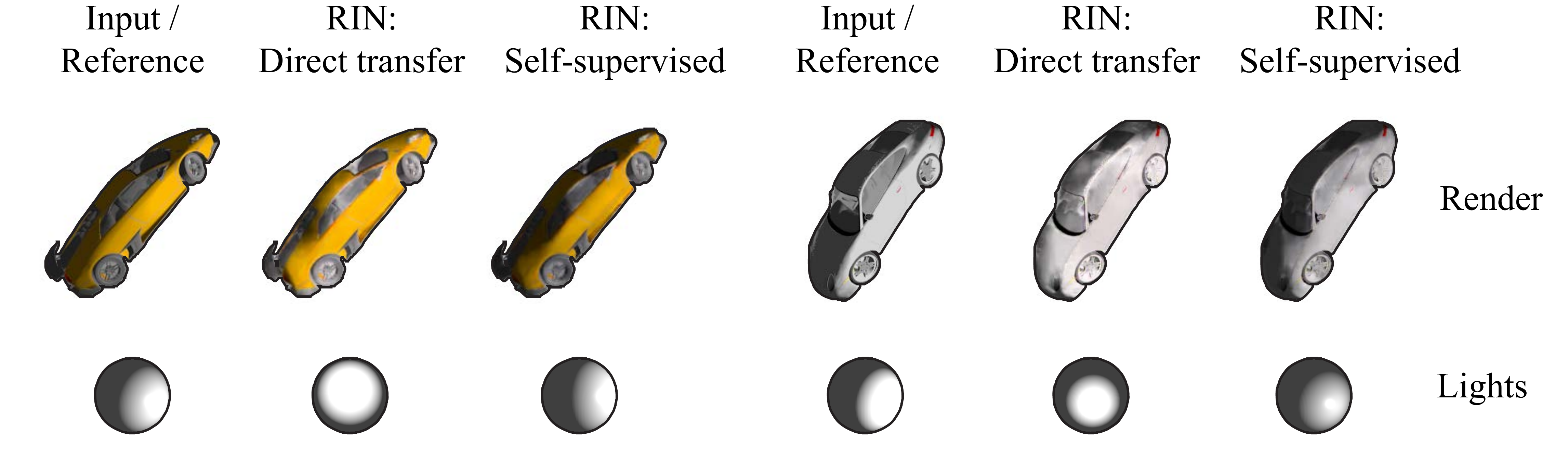}
    \caption{Predictions of RIN trained on left-lit images before and after self-supervised learning on right-lit images. RIN uncovers the updated lighting distribution without external supervision or ground truth data.}
    \label{fig:gen_lights}
\end{figure}

\subsection{Lighting transfer}

\textbf{Data} \hspace{0.35cm} Cars from the ShapeNet 3D model repository were rendered at random orientations and scales. In the labeled data, they were lit only from the left side. In the unlabeled data, they were lit from both the left and right.

\textbf{Results} \hspace{0.35cm} Before self-supervised training on the unlabeled data, the model's distribution over lighting predictions mirrored that of the labeled training set. When tested on images lit from the right, then, it tended to predict centered lighting. After updating the lighting decoder based on reconstruction error from these right-lit images though, the model's lighting predictions more accurately reflected the new distribution and lighting mean-squared error reduces by 18\%. Lighting predictions, along with reconstructions, for right-lit images are shown in Fig~\ref{fig:gen_lights}.

\begin{figure}[t]
    \centering
    \includegraphics[width=\linewidth]{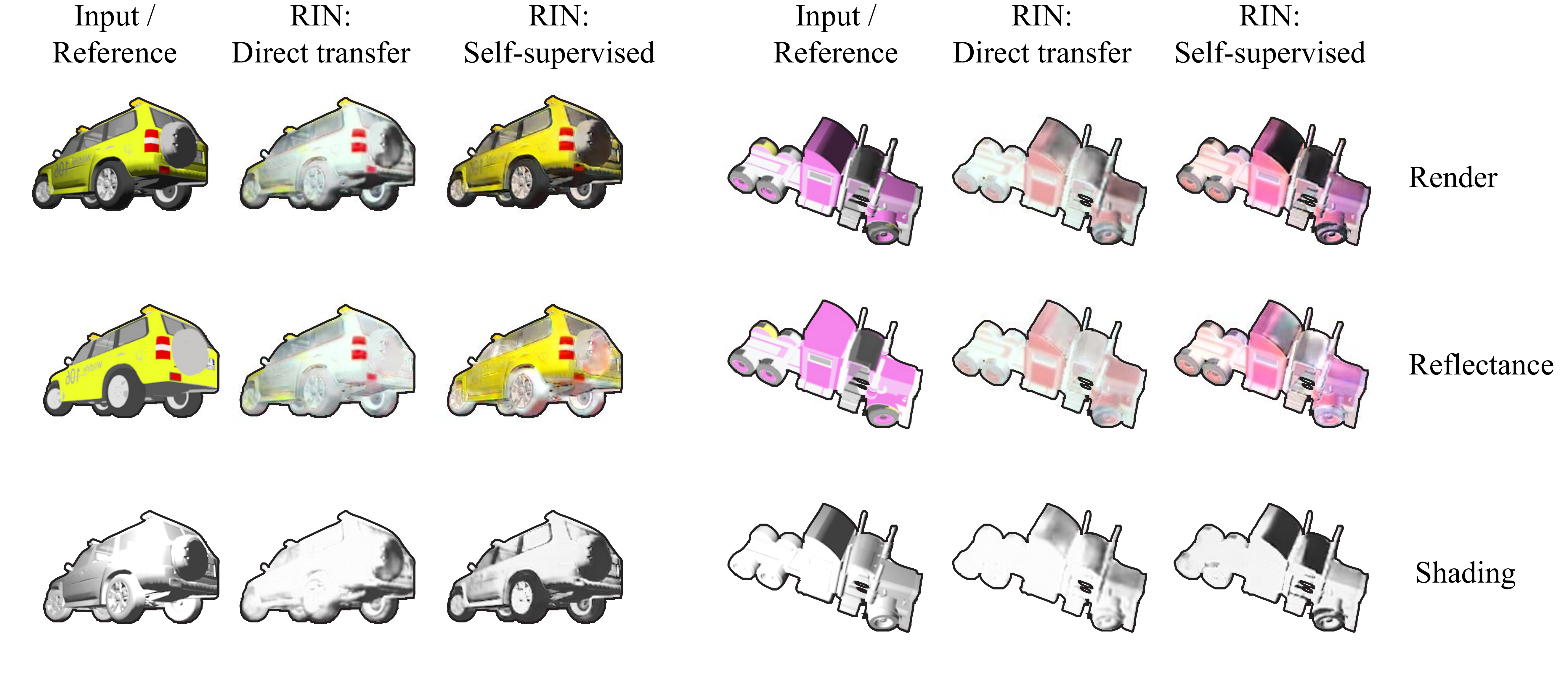}
    \caption{RIN was first trained on ShapeNet airplanes and then tested on cars. Because most of the airplanes were white, the reflectance predictions were washed out even for colorful cars. RIN fixed the mismatch between datasets without any ground truth intrinsic images of cars.}
    \label{fig:gen_category}
\end{figure}

\subsection{Category transfer}
In the previous transfer experiments, only one intrinsic image was mismatched between the labeled and unlabeled data, so only one of RIN's decoders needed updating during transfer. When transferring between object categories, though, there is not such a guarantee. Although it might be expected that a model trained on sufficiently many object categories would learn a generally-useful distribution over reflectances, it is difficult to ensure that this is the case. We are interested in these sorts of scenarios to determine how well self-supervised transfer works when more than one decoder needs to be updated to account for unlabeled data.

\textbf{Data} \hspace{0.35cm} Datasets of ShapeNet cars and airplanes were created analogously to those in Section~\ref{sec:supervised}. The airplanes had a completely different color distribution than the cars as they were mostly white, whereas the cars had a more varied reflectance distribution. The airplanes were used as the labeled category to ensure a mismatch between the train and transfer data. 

\textbf{Results} \hspace{0.35cm} To transfer to the new category, we allowed updates to all three of the RIN decoders. (The shader was left fixed as usual.) There were pronounced improvements in the shading predictions (32\%) accompanied by modest improvements in reflectances (21\%). The shading predictions were not always caused by improved shape estimates. Because there is a many-to-one mapping from shape to shading (conditioned on a lighting condition), it is possible for the shape predictions to worsen in order to improve the shading estimates. The lighting predictions also remained largely unchanged, although for the opposite reason: because no lighting region were intentionally left out of the training data, the lighting predictions were adequate on the transfer classes even without self-supervised learning. 

\begin{table}[t]
    \centering
    \begin{tabular}{lcccccccccc}
        \toprule
        & Reflectance & Shape & Lights & Shading & Render \\
        \midrule
        Direct transfer & 0.019 & 0.014 & 0.584 & 0.065 & 0.035 \\
        Self-supervised & 0.015 & 0.014 & 0.572 & 0.044 & 0.006 \\
        \bottomrule
    \end{tabular}
    \label{tbl:gen_category}
    \vspace{.5cm}
    \caption{MSE of RIN trained on ShapeNet airplanes before and after self-supervised transfer to cars. Although RIN improves its shading predictions, these are not necessarily driven by an improvement in shape prediction.}
    \vspace{-20pt}
\end{table}

%% file: text/conclusion.tex
\section{Conclusion}
In this paper, we proposed the Rendered Intrinsics Network for intrinsic image prediction. We showed that by learning both the image decomposition and recombination functions, RIN can make use of reconstruction loss to improve its intermediate representations. This allowed unlabeled data to be used during training, which we demonstrated with a variety of transfer tasks driven solely by self-supervision. When there existed a mismatch between the underlying intrinsic images of the labeled and unlabeled data, RIN could also adapt its predictions in order to better explain the unlabeled examples. 







%% file: ms.bbl
\begin{thebibliography}{30}
\providecommand{\natexlab}[1]{#1}
\providecommand{\url}[1]{\texttt{#1}}
\expandafter\ifx\csname urlstyle\endcsname\relax
  \providecommand{\doi}[1]{doi: #1}\else
  \providecommand{\doi}{doi: \begingroup \urlstyle{rm}\Url}\fi

\bibitem[Barron and Malik(2015)]{sirfs}
Jonathan~T Barron and Jitendra Malik.
\newblock Shape, illumination, and reflectance from shading.
\newblock \emph{IEEE TPAMI}, 37\penalty0 (8):\penalty0 1670--1687, 2015.

\bibitem[Barrow and Tenenbaum(1978)]{barrow_tenenbaum}
H.G. Barrow and J.M. Tenenbaum.
\newblock Recovering intrinsic scene characteristics from images.
\newblock \emph{Computer Vision Systems}, 1978.

\bibitem[Bell et~al.(2014)Bell, Bala, and Snavely]{crowdsourced}
Sean Bell, Kavita Bala, and Noah Snavely.
\newblock Intrinsic images in the wild.
\newblock \emph{ACM TOG}, 33\penalty0 (4):\penalty0 159, 2014.

\bibitem[Chang et~al.(2015)Chang, Funkhouser, Guibas, Hanrahan, Huang, Li,
  Savarese, Savva, Song, Su, et~al.]{shapenet}
Angel~X Chang, Thomas Funkhouser, Leonidas Guibas, Pat Hanrahan, Qixing Huang,
  Zimo Li, Silvio Savarese, Manolis Savva, Shuran Song, Hao Su, et~al.
\newblock Shapenet: An information-rich 3d model repository.
\newblock \emph{arXiv preprint arXiv:1512.03012}, 2015.

\bibitem[Chen et~al.(2016)Chen, Chen, Duan, Houthooft, Schulman, Sutskever, and
  Abbeel]{infogan}
Xi~Chen, Xi~Chen, Yan Duan, Rein Houthooft, John Schulman, Ilya Sutskever, and
  Pieter Abbeel.
\newblock Infogan: Interpretable representation learning by information
  maximizing generative adversarial nets.
\newblock In \emph{NIPS}, 2016.

\bibitem[Grosse et~al.(2009)Grosse, Johnson, Adelson, and
  Freeman]{mit_intrinsic}
Roger Grosse, Micah~K. Johnson, Edward~H. Adelson, and William~T. Freeman.
\newblock Ground-truth dataset and baseline evaluations for intrinsic image
  algorithms.
\newblock In \emph{ICCV}, 2009.

\bibitem[He et~al.(2016)He, Zhang, Ren, and Sun]{residual_learning}
Kaiming He, Xiangyu Zhang, Shaoqing Ren, and Jian Sun.
\newblock Deep residual learning for image recognition.
\newblock In \emph{CVPR}, 2016.

\bibitem[Hinton et~al.(2011)Hinton, Krizhevsky, and Wang]{hinton_autoencoder}
Geoffrey~E Hinton, Alex Krizhevsky, and Sida~D Wang.
\newblock Transforming auto-encoders.
\newblock In \emph{ICANN}, 2011.

\bibitem[Hold-Geoffroy et~al.(2017)Hold-Geoffroy, Sunkavalli, Hadap,
  Gambaretto, and Lalonde]{deep_illumination}
Yannick Hold-Geoffroy, Kalyan Sunkavalli, Sunil Hadap, Emiliano Gambaretto, and
  Jean-Francois Lalonde.
\newblock Deep outdoor illumination estimation.
\newblock In \emph{CVPR}, 2017.

\bibitem[Horn(1974)]{horn}
Berthold~K.P. Horn.
\newblock Determining lightness from an image.
\newblock \emph{Computer Graphics and Image Processing}, 3:\penalty0 277--299,
  1974.

\bibitem[Innamorati et~al.(2017)Innamorati, Ritschel, Weyrich, and
  J.~Mitra]{innamorati}
Carlo Innamorati, Tobias Ritschel, Tim Weyrich, and Niloy J.~Mitra.
\newblock Decomposing single images for layered photo retouching.
\newblock \emph{Computer Graphics Forum}, 36:\penalty0 15--25, 07 2017.

\bibitem[Ioffe and Szegedy(2015)]{batchnorm}
Sergey Ioffe and Christian Szegedy.
\newblock Batch normalization: Accelerating deep network training by reducing
  internal covariate shift.
\newblock In \emph{ICML}, 2015.

\bibitem[Kar et~al.(2015)Kar, Tulsiani, Carreira, and Malik]{category_specific}
Abhishek Kar, Shubham Tulsiani, Joao Carreira, and Jitendra Malik.
\newblock Category-specific object reconstruction from a single image.
\newblock In \emph{CVPR}, 2015.

\bibitem[Kingma and Ba(2015)]{adam}
Diederik~P. Kingma and Jimmy Ba.
\newblock Adam: A method for stochastic optimization.
\newblock In \emph{ICLR}, 2015.

\bibitem[Kulkarni et~al.(2015)Kulkarni, Whitney, Kohli, and Tenenbaum]{dcign}
Tejas~D Kulkarni, William~F Whitney, Pushmeet Kohli, and Josh Tenenbaum.
\newblock Deep convolutional inverse graphics network.
\newblock In \emph{NIPS}, 2015.

\bibitem[Land and McCann(1971)]{retinex}
Edwin~H. Land and John~J. McCann.
\newblock Lightness and retinex theory.
\newblock \emph{Journal of the Optical Society of America}, 61:\penalty0 1--11,
  1971.

\bibitem[LeCun et~al.(2015)LeCun, Bengio, and Hinton]{lecun2015deep}
Yann LeCun, Yoshua Bengio, and Geoffrey Hinton.
\newblock Deep learning.
\newblock \emph{Nature}, 521\penalty0 (7553):\penalty0 436--444, 2015.

\bibitem[Lombardi and Nishino(2012)]{lombardi_cvpr_2012}
Stephen Lombardi and Ko~Nishino.
\newblock Single image multimaterial estimation.
\newblock In \emph{CVPR}, 2012.

\bibitem[Lombardi and Nishino(2016)]{lombardi_tpami_2016}
Stephen Lombardi and Ko~Nishino.
\newblock Reflectance and illumination recovery in the wild.
\newblock \emph{IEEE TPAMI}, 38\penalty0 (1):\penalty0 129--141, 2016.

\bibitem[Nalbach et~al.(2017)Nalbach, Arabadzhiyska, Mehta, Seidel, and
  Ritschel]{deep_shading}
Oliver Nalbach, Elena Arabadzhiyska, Dushyant Mehta, Hans-Peter Seidel, and
  Tobias Ritschel.
\newblock Deep shading: Convolutional neural networks for screen-space shading.
\newblock \emph{Computer Graphics Forum}, 36\penalty0 (4), 2017.

\bibitem[Narihira et~al.(2015{\natexlab{a}})Narihira, Maire, and
  Yu]{direct_intrinsics}
Takuya Narihira, Michael Maire, and Stella~X. Yu.
\newblock Direct intrinsics: Learning albedo-shading decomposition by
  convolutional regression.
\newblock In \emph{ICCV}, 2015{\natexlab{a}}.

\bibitem[Narihira et~al.(2015{\natexlab{b}})Narihira, Maire, and
  Yu]{narihari_lightness}
Takuya Narihira, Michael Maire, and Stella~X. Yu.
\newblock Learning lightness from human judgement on relative reflectance.
\newblock In \emph{CVPR}, 2015{\natexlab{b}}.

\bibitem[Oxholm and Nishino(2016)]{oxholm_tpami_2016}
Geoffrey Oxholm and Ko~Nishino.
\newblock Shape and reflectance estimation in the wild.
\newblock \emph{IEEE TPAMI}, 38\penalty0 (2):\penalty0 376--389, 2016.

\bibitem[Qu{\'e}au and Durou(2015)]{beethoven}
Yvain Qu{\'e}au and Jean-Denis Durou.
\newblock Edge-preserving integration of a normal field: Weighted
  least-squares, tv and {L}1 approaches.
\newblock In \emph{International Conference on Scale Space and Variational
  Methods in Computer Vision}, 2015.

\bibitem[Rematas et~al.(2016)Rematas, Ritschel, Fritz, Gavves, and
  Tuytelaars]{deep_reflectance}
Konstantinos Rematas, Tobias Ritschel, Mario Fritz, Efstratios Gavves, and
  Tinne Tuytelaars.
\newblock Deep reflectance maps.
\newblock In \emph{CVPR}, June 2016.

\bibitem[Shi et~al.(2017)Shi, Dong, Su, and Yu]{non_lambertian}
Jian Shi, Yue Dong, Hao Su, and Stella~X. Yu.
\newblock Learning non-lambertian object intrinsics across shapenet categories.
\newblock In \emph{CVPR}, 2017.

\bibitem[Shu et~al.(2017)Shu, Yumer, Hadap, Sunkavalli, Shechtman, and
  Samaras]{shu2017face}
Zhixin Shu, Ersin Yumer, Sunil Hadap, Kalyan Sunkavalli, Eli Shechtman, and
  Dimitris Samaras.
\newblock Neural face editing with intrinsic image disentangling.
\newblock In \emph{CVPR}, July 2017.

\bibitem[Tang et~al.(2012)Tang, Salakhutdinov, and Hinton]{dln}
Yichuan Tang, Ruslan Salakhutdinov, and Geoffrey Hinton.
\newblock Deep lambertian networks.
\newblock In \emph{ICML}, 2012.

\bibitem[Weiss(2001)]{sequences}
Yair Weiss.
\newblock Deriving intrinsic images from image sequences.
\newblock In \emph{ICCV}, 2001.

\bibitem[Wu et~al.(2017)Wu, Wang, Xue, Sun, Freeman, and Tenenbaum]{marrnet}
Jiajun Wu, Yifan Wang, Tianfan Xue, Xingyuan Sun, William~T Freeman, and
  Joshua~B Tenenbaum.
\newblock Marrnet: 3d shape reconstruction via 2.5d sketches.
\newblock In \emph{NIPS}, 2017.

\end{thebibliography}
